%% file: 0_root.tex
\documentclass[letterpaper, 10 pt, conference]{ieeeconf}  

\IEEEoverridecommandlockouts                              
\usepackage{times}
\usepackage{multicol}
\usepackage[bookmarks=true]{hyperref}
\usepackage{graphicx}
\usepackage{subcaption}
\usepackage{xcolor}
\usepackage[linesnumbered,ruled,vlined]{algorithm2e}
\usepackage{diagbox}
\usepackage{amsmath}
\usepackage{amssymb}
\usepackage{dblfloatfix}
\usepackage{siunitx}  
\usepackage[ruled,vlined]{algorithm2e}
\title{\LARGE \bf Measuring Uncertainty in Shape Completion to Improve Grasp Quality}

\author{Nuno Ferreira Duarte$^{1}$ \and Seyed S. Mohammadi$^{3}$ \and Plinio Moreno$^{1}$ \and Alessio {Del Bue}$^{2}$ \and José Santos-Victor$^{1}$ 
\thanks{*This work has partially received funding from the European Union’s Horizon 2020 research and innovation programme under grant agreement No 964854; the FCT funding to the ISR/LARSyS Associated Laboratory UID/EEA/50009/2020 and LA/P/0083/2020 
}
\thanks{$^{1}$Vislab, ISR-Lisboa, Instituto Superior T\'{e}cnico, Universidade de Lisboa, Portugal. Email:{\tt\small$\{$nferreiraduarte, plinio, jasv$\}$@isr.tecnico.ulisboa.pt}} %
\thanks{$^{2}$PAVIS, Fondazione Istituto Italiano di Tecnologia, Genoa, Italy.}
\thanks{$^{3}$Leonardo, Genoa, Italy.}
}

\begin{document} 

\maketitle
\thispagestyle{empty}
\pagestyle{empty}

\begin{abstract}
Shape completion networks have been used recently in real-world robotic experiments to complete the missing/hidden information in environments where objects are only observed in one or few instances where self-occlusions are bound to occur. Nowadays, most approaches rely on deep neural networks that handle rich 3D point cloud data that lead to more precise and realistic object geometries. However, these models still suffer from inaccuracies due to its nondeterministic/stochastic inferences which could lead to poor performance in grasping scenarios where these errors compound to unsuccessful grasps.
We present an approach to calculate the uncertainty of a 3D shape completion model during inference of single view point clouds of an object on a table top. In addition, we propose an update to grasp pose algorithms quality score by introducing the uncertainty of the completed point cloud present in the grasp candidates. To test our full pipeline we perform real world grasping with a 7dof robotic arm with a 2 finger gripper on a large set of household objects and compare against previous approaches that do not measure uncertainty. Our approach ranks the grasp quality better, leading to higher grasp success rate for the rank 5 grasp candidates compared to state of the art. 

\end{abstract}

\input{1_intro.tex}
\input{2_model_uncertainty.tex}

\input{3_grasp_score.tex}

\input{4_experiments.tex}
\input{5_conclusion.tex}


\bibliographystyle{ieeetr}
\bibliography{IEEEabrv, refs}

\end{document}

%% file: 1_intro.tex
\section{Introduction}

In the last decades grasping has been an on going challenge in robotics, and although much progress has occurred since the very first attempts, it has yet to reach a mature and robust state for all grasping scenarios and object types.  
Recently, with the desire of having robots working along side humans there is a need for making robots adaptable to ever changing environments and objects. With this necessity comes the challenge of grasping new or not seen before objects which requires sensors such as cameras. The problem arises when these sensors can not view the full shape of the object for a proper grasp without collisions or failure. One of the solutions for enhancing the information of the camera sensors is to use prior knowledge of previously observed objects to fill the missing information. These solutions are called shape completion and process RGB or RGBD information as a voxel grid or point cloud data to generate the occluded region of the object. These approaches improve significantly the grasp successfulness and has been used recently for many tasks \cite{han2017high, mohammadi2022svp, varley2017shape}. 

Although these approaches assist the grasping of objects in real world scenarios and some solutions allow for online inference of shape completion models and generation of grasping candidates, it does not come without its limitations. The sensors have inherent limitations and the shape completion model inference have its  own noise in the generated output. Some the inference noise could result from the camera sensor noise, low resolution, or light exposure. These errors are on the hardware side and can be mitigated by more reliable (expensive) cameras. However, another noise comes from the model itself and it should be accounted for when selecting which grasps are best for a successful attempt. The latter is the focus of this work and our proposal is aimed to tackle such noise. 

\begin{figure}[t!]
  \centering
  \includegraphics[width=0.5\textwidth]{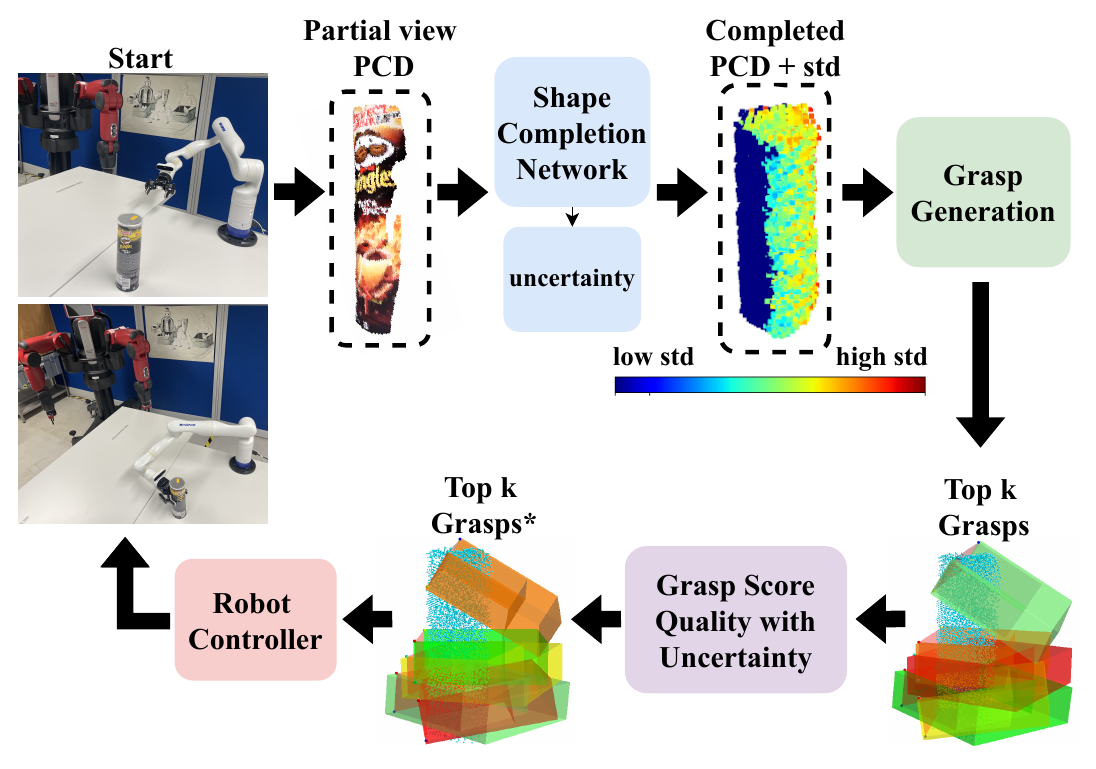}
  \caption{The proposed approach for measuring uncertainty in the shape completion models and including uncertainty in grasp generator algorithms. Our model receives the partial PCD and generates a completed PCD for the object shape and the PCD's std corresponding low/high std as low/high uncertainty points, respectively. The default grasp algorithm generates grasp candidates and rank it using a grasp score metric. We proposed an extension to include the uncertainty in the grasp score metric. *The same grasp candidates have a new ranking according to their new grasp quality score.}
  \label{fig:pipeline}
\end{figure}

In our work we propose a new ranking for grasps generated from object shape-completion models using models' completion uncertainty. The procedure, shown in Figure \ref{fig:pipeline},  starts with a camera, attached to the robotic gripper, taking a single view of an object on the table and acquiring the PCD. In the next step, the PCD is segmented to remove the table, leaving only the visible (partial PCD) part of the object. 
We include our proposed uncertainty layer in a state-of-the-art shape completion network \cite{mohammadi_3dsgrasp_2023} to complete the shape (completed PCD) and the associated noise of the completed PCD. 
Afterwards, we proceed to generate grasp candidates from an off-the-shelf algorithm \cite{pas_grasp_2017} that gives us the 15 best grasp candidates ranked with the corresponding grasp quality scores. Our next approach then uses the uncertainty of the completed PCD to update the grasp quality score of all 15 grasps candidates. This process involves extracting just the PCD inside each grasp pose candidate (cropped PCD) and adding the cropped uncertainty to the respective score. Following the updated grasp scores, the 15 grasps candidates are reordered according to the new ranking of quality score. Finally, the new top 5 grasps are executed on the robot to evaluate the performance of successfully grasping with our approach. We evaluate our approach against the 3DSGrasp pipeline which executes the top 5 grasps generated directly from \cite{pas_grasp_2017}. Our solution provides an improvement in the grasp success rate in YCB objects proving the advantage of accounting for the noise in the grasp quality score.

Our contributions can then be summarized to: 
\begin{itemize}
    \item Compute PCD shape-completion uncertainty

    \item Propose grasp quality score which accounts for PCD shape-completion uncertainty

\end{itemize}

The structure of the paper is as follows: Section \ref{sec:soa} is reserved for literature review, in Section \ref{sec:model} the uncertainty layer is introduced, in Section \ref{sec:grasp} we present the uncertainty measurement as an extension to the grasp candidates scores, the Section \ref{sec:robot} shows the robot setup and results of our grasping approach and other grasping methods. The paper finishes with discussion on the performance of the different methods tested, main conclusions and further extensions proposed as future work.

\section{Related Work}

\label{sec:soa}
This sections addresses the current state of the art for robotic grasping using visual information with an emphasis on 3D data as input. It also presents the current trends for computing and mitigating the uncertainty problem in grasping under real robot scenarios. 

\noindent \textbf{Robot grasping using object shape completion.} 
Contrary to most grasp approaches, it is not always possible to grasp every object using a top-down grasp \cite{fang2020graspnet, fang_anygrasp_2023}. In human-shared environments, the robot will not have a good view of the object without occlusions, hence models for shape completion provide crucial information for better grasps. These methods can be learned using prior geometric knowledge \cite{berger2014state,de2015efficient, kazhdan2006poisson}, considered as model-based methods, or more recently, data-driven methods with deep learning techniques that have allowed to recreate with high accuracy the shape of a wide variety of objects and handle richer 3D information \cite{gurumurthy2019high, han2017high}.
The early attempts to process higher dimensional data, point cloud data, was to convert to a more conventional data representation, by voxelization, which would then be processed by deep convolutional neural networks \cite{han2017high, mohammadi2022svp, varley2017shape}. Nowadays, with the introduction of the PointNet architecture \cite{qi2017pointnet} and with higher computational resources, it is possible to compute raw PCD for classification tasks. It then followed to object shape completion tasks like PCN \cite{yuan2018pcn} and FoldingNet \cite{yang2018foldingnet}. Newer learning-based  architectures then surpassed them for 3D shape completion tasks using PCD input \cite{groueix2018papier, liu2020morphing, pan2021multi, yu2021PoinTr} which improved the resolution and accuracy of the reconstructed shape, while others \cite{chen2022improving,PointSDFgrasping2020, yang2021robotic, humt_combining_2023} improve the grasp success rate of the objects grasped. The introduction of Transformer networks arose the use of transformer backbones with PCD input \cite{chen2022improving, mohammadi_3dsgrasp_2023, yu2021PoinTr} with increase in performance on 3D PCD shape completion. 3DSGrasp \cite{mohammadi_3dsgrasp_2023} improves the reconstruction of object's shapes compared to the state of the art and provide more reliable grasp candidates. 

\noindent \textbf{Vision-based grasping uncertainty-aware.} 
There have been works that explore the uncertainty present in real context scenarios such as grasping unknown objects using vision.  
Li et al. \cite{li2016dexterous}, Rosasco et al. \cite{rosasco2022towards}, and Chen et al. \cite{chen2018probabilistic} have used Gaussian processes, implicit function, and signed distance function, respectively, to complete the shape of the partial image of the object from the camera. These methods can calculate the shape uncertainty to provide the preferred contact points for a safe grasp. However these shape completion methods can mostly reconstruct smooth surfaces which results in low details on the real object shape. Lundell et al. \cite{lundell2019robust} proposed Monte-Carlo dropout to compute uncertainty for shape completion. Their approach reconstructs the shape as a voxel grid and takes the average shape completion to compute the grasps samples. Our approach uses Monte-Carlo dropout on a point-based network which provides better resolution to reconstruct the shape, and in addition, we use the average shape and the uncertainty computed to generate and qualify the grasps. 
More recently, Rustler et al. \cite{rustler_active_2022, rustler_efficient_2023} has proposed a shape completion model using implicit surface networks which produce reconstructed shape with higher quality. In both works they measure uncertainty using a vision-haptic-based system that requires the robot to actively touch the object in the regions of highest uncertainty to update the current object's shape. This approach might not be feasible for all situations given that touching an object may lead to unforeseen collisions which could result in unrecoverable failures. Humt et al. \cite{humt2023shape} have also proposed a shape completion model that can predict regions of uncertainty, however, this model is used mainly for predicting the unseen region of mugs in order to identify the handle portion. The goal is to reconstruct mugs and if the handle is not visible by the camera, it adds an uncertainty region so that the grasp contact avoids that space. 

\noindent \textbf{Grasping Evaluation.} To assess the performance of grasping approaches, the most common approach is to compute the success on the pose with largest score (i.e. top 1) . However, to have a better notion of the grasps and evaluate the grasp pose distribution, the Graspnet-1billion \cite{fang2020graspnet} uses the $Precision@k$. In practice, it is estimated empirically on unseen objects by executing the top k grasps and checking the grasp successes. We follow this approach, computing the common $Precision@1$ and $Precision@5$.

%% file: 2_model_uncertainty.tex
\section{Shape Completion Uncertainty}
\label{sec:model}
In this section we introduce our proposed uncertainty aware 3D shape completion network. 

\subsection{Shape Completion}
For the shape completion network we use the state-of-the-art approach on household objects, 3DSGrasp \cite{mohammadi_3dsgrasp_2023}, which was trained on the YCB dataset and achieved the best accuracy for robotic grasping. 3DSGrasp is a transformer architecture that contains a multi-head Offset-Attention encoder-decoder Transformer layer \cite{guo2021pct} which takes as a input the embedded shape feature and output predicted shape features representing the feature vector of the missing point cloud. The point cloud generation network will predict the missing geometry part of the completed data by passing the predicted shape feature through FoldingNet \cite{yang2018foldingnet}. Finally, the generated point cloud will merge with the input partial data to shape the completed shape object.


\begin{figure}[h]
    \centering
    \begin{subfigure}{.49\textwidth}
      \centering
      \includegraphics[width=.99\linewidth]{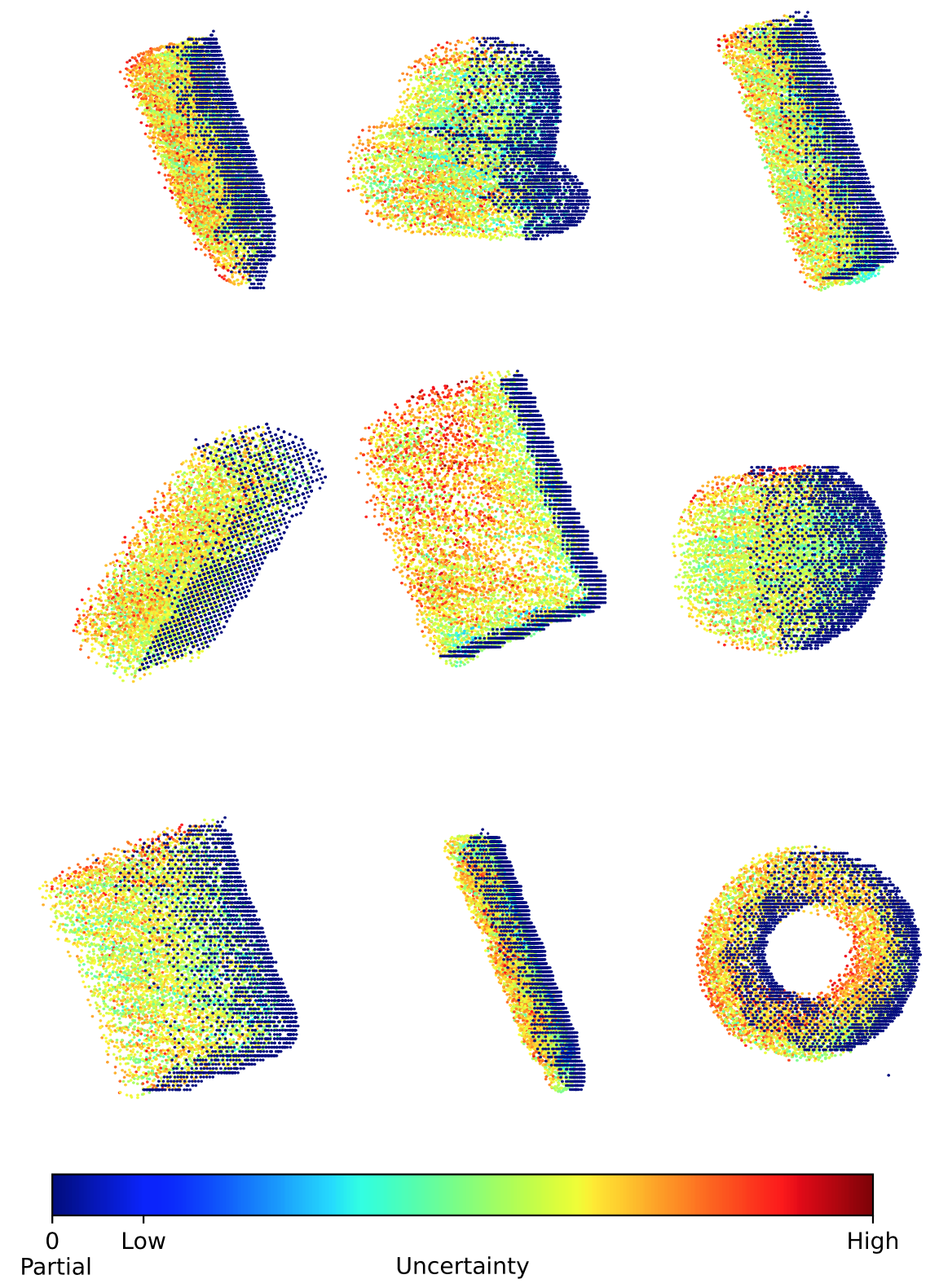}
    \end{subfigure}%
     \caption{Shape completion and respective uncertainty values for different partial PCD of objects present in the YCB dataset. These partials are segments including only surface points of the YCB object dataset. Top to bottom, left to right: french yellow mustard, mushroom, pringles original, black and decker lithium drill box, box, mini soccer ball, tomato soup, detergent bottle, donut.}
    \label{fig:ycb_std}
\end{figure}

\subsection{Monte Carlo Dropout as uncertainty estimation}

The dropout is one of the most fundamental regularization techniques in machine and deep  learning, which applies to improve model training. Its main objective is to optimize the adjusted loss function while avoiding overfitting and underfitting. The dropout rate is an important parameter in this technique, it typically ranges from 0 (i.e., no dropout) to 0.5 (i.e., approximately half of all neurons being deactivated). The selection of the specific dropout rate is contingent upon various factors such as the architecture of the network and the dimensions of its layers. This dynamic dropout process contributes to the regularization of the model and facilitates its adaptability across different training scenarios.
The Monte Carlo Dropout technique introduced by \cite{pmlr-v48-gal16}, estimates the uncertainty of the predictions of a Neural Network architecture. The main assumption is to apply dropout during the training process, to be able to approximate the distribution over weight matrices: $q(\boldsymbol{\omega})$. For a new sample point $\boldsymbol{x^*}$, the uncertainty is estimated as follows~\cite{pmlr-v48-gal16}:
\begin{align}
    \mathbb{E}_{q(\boldsymbol{y^*}|\boldsymbol{x^*})}(\boldsymbol{y^*}) \approx \frac{1}{T}\sum_{t=1}^{T} \boldsymbol{\hat{y}^*}(\boldsymbol{x^*},\boldsymbol{W}_1^t,\cdots,\boldsymbol{W}_L^t) \label{eqn:uncertainty},
\end{align}
where $\boldsymbol{W}_i$ is the weight matrix with zeros on deactivated connections at layer $i$, $L$ corresponds to the number of layers, and $T$ to the number of forward passes. After the training process with dropout, the approximation in Equation \eqref{eqn:uncertainty} corresponds to $T$ forward passes, which is set to $T=60$ for our experiments. The output distribution not only encapsulates the model's mean prediction but also encapsulates a measure of uncertainty in the predictive outcomes. As a result, the utilisation of Monte Carlo Dropout serves the dual purpose of refining predictions while offering a systematic means of regularising the network by incorporating uncertainty considerations into the model's architecture.

\subsection{Uncertainty of Objects' Shape}
This section illustrates the integration of the Monte Carlo Dropout with the point cloud completion network 3DSGrasp \cite{mohammadi_3dsgrasp_2023}. In each iteration during training, a random subset of neurons is dropped out, introducing variability in the model parameters. For the inference time, the traditional dropout is turned off, leading to fixed predictions, while in Monte Carlo Dropout, dropout is retained during the inference time as well. We also run the model through multiple forward passes during inference, where in each forward pass, dropout is applied, resulting in different predictions for the same input. The predictions from these multiple forward passes are then aggregated to obtain a final prediction. The completed PCD is set as the average point positions from all the multiple inferences. We then estimate the uncertainty by calculating the standard deviation or variance of the prediction across multiple passes. Figure \ref{fig:ycb_std} shows the result of the completion network with the uncertainty layer. The points are colored according to the standard deviation value where blue values represent low uncertainty and on the opposite side of the spectrum the red values are for high uncertainty. The values for standard deviation range from 2mm to 6mm. It can be seen from the different object reconstructions that the high uncertainty values are usually located in the far side, i.e. the furthest points from the partial PCD.

%% file: 3_grasp_score.tex
\section{Grasping Score}
\label{sec:grasp}

In this section we present the step-by-step procedure to compute the uncertainty provided by our approach in Section \ref{sec:model} to any grasp generation algorithm that uses 3D point clouds. For the purpose of this work we went with the GPD \cite{pas_grasp_2017} but any grasp pose detector/generator would function similarly.

\begin{figure*}[h]
    \centering
    \includegraphics[width=0.95\textwidth]{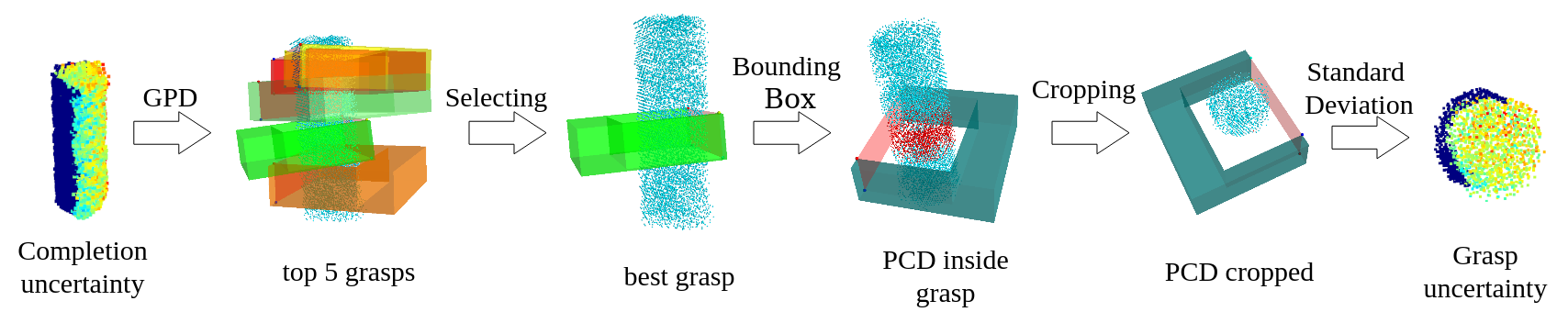}
    \caption{Adding the uncertainty in the grasp quality score of one grasp candidate (best ranked grasp) for the complete PCD of the pringles can. The algorithm iterates from left to right: (i) output of the shape completion network with uncertainty layer; (ii) output of GPD \cite{pas_grasp_2017} for the complete PCD; (iii) for each grasp, start with the first; (iv) find the 3D points in the complete PCD that are inside the gripper; (v) crop the 3D points and get the standard deviation of the points in the cropped PCD; (vi) add the uncertainty of those points to the grasp quality score.}
    \label{fig:gpd_compute_std}
\end{figure*}

The scoring of uncertainty in the grasp pose candidates is the main reason for measuring the shape completion standard deviation. In section \ref{sec:model} the shape completion model infers multiple times the same partial PCD which results in the completed PCD as the average of those multiple inferences and the uncertainty as the standard deviation. The state of the art 3D point cloud grasp candidates generators only receive the 3D partial/complete PCD, no noise or uncertainty variables can be included, hence by running the off-the-shelf algorithms would only result in grasp scores of the average complete PCD. Our proposal extends the algorithm by including the standard deviation of the complete PCD in the grasp candidates scores which updates the quality output resulting in a new ranked order for the quality grasp score. Figure \ref{fig:gpd_compute_std} illustrates our proposed steps for computing the uncertainty for each grasp candidate generated by GPD \cite{pas_grasp_2017} of the complete PCD from the shape completion network in Section \ref{sec:model}.

\begin{algorithm}[h]
\DontPrintSemicolon
  
  \textbf{Input} {completed PCD ($CP$), std of CP ($CP_{std}$)} \\
  \textbf{Output} {a set of 6-DOF grasp candidates $G^*$ ranked according to grasp score $S'$}

  $G$, $S$ = GraspGeneration($CP$) 
  
  $C$ = ContactPoints($G$) 
  
  $BBox$ = BoudingBox($C$, $CP$) 
  
  $CP'$, $CP'_{std}$ = Crop($BBox$, $CP$) 
  
  $S'$ = Score($S$, $CP'$, $CP'_{std}$) 
  
  $G^*$ = Order($G$, $S'$) 

\caption{Grasp Score with Shape Uncertainty}
\label{alg:grasp_alg}
\end{algorithm}

The step-by-step process to update the grasp score is shown in Algorithm \ref{alg:grasp_alg}. 
Step 1 runs grasp generation (GPD \cite{pas_grasp_2017} or PointNetGPD \cite{liang_2019_pointnetgpd}) inference on the complete PCD and saves the grasp pose candidates and the corresponding scores. Step 2 takes each grasp pose and gets the contact points of the gripper when applying the grasp pose on the completed PCD. Step
3 creates a bounding box with the contact points that encapsulates the PCD of the completed PCD. Step 4 crops the completed PCD from the bounding box. Step 5 computes the new grasp score for each grasp pose by including the uncertainty of the cropped section of the completed PCD. Step 6 reorders the grasp scores from best to worst. 

\begin{figure}
    \centering
    \begin{subfigure}{.25\textwidth}
      \centering
      \includegraphics[width=.99\linewidth]{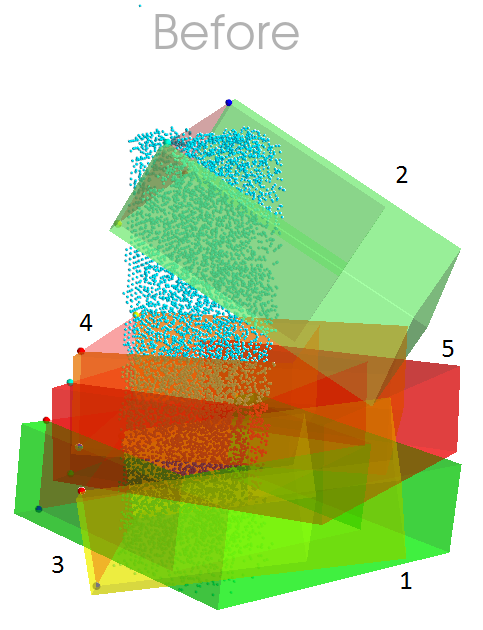}
    \end{subfigure}%
    \begin{subfigure}{.25\textwidth}
      \centering
      \includegraphics[width=.99\linewidth]{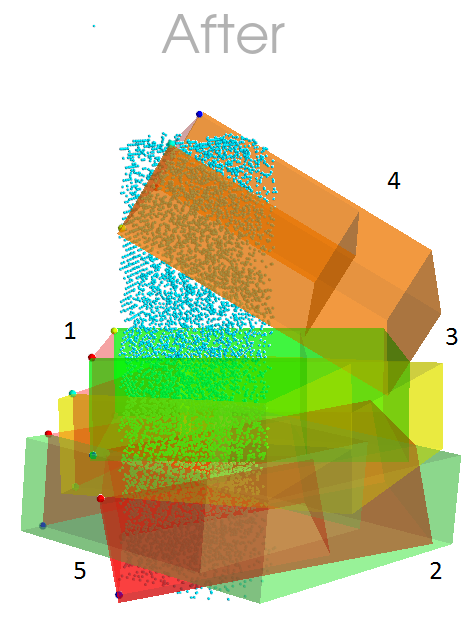}
    \end{subfigure}
    \caption{An example of our proposed grasp quality score showing the before and after output for the completed PCD of the pringles can. Before is the GPD output and after is our proposed algorithm. The after ranking index is [4, 1, 5, 2, 3] for the top 5 grasp pose candidates. Colors from best to worst: bright green, light green, yellow, orange, red.}
    \label{fig:before_after_std}
\end{figure}

In step 1 a list of 6DoF grasp pose candidates for a two-fingered gripper are generated. In our work we use the 2f-85 robotiq gripper in our robotic setup. 
For step 2, given that we know the gripper dimensions, we can extract the contact points of the gripper. We use the dex-net library tools \cite{liang_2019_pointnetgpd} to extract the 3D contact points of the gripper for each grasp pose candidate. These gives the 8 3D points corresponding to the gripper's two inner surface points, one for each of the parallel fingers. 
In step 3 those contact points are used to define the bounding box representing the volume that gets inside the gripper's fingers. The bounding box then extracts the 3D points of the completed PCD
which is used in step 4 to crop the PCD to multiple cropped PCDs, one for each grasp pose candidate. During this step, the standard deviation of the points is also segmented to include the uncertainty of only the cropped points. This finishes with each grasp pose candidate having a cropped PCD version and the associated point's standard deviation. 
The final two steps, relate to updating the grasp quality score. In step 5, we extend the score function, shown in Equation \ref{eq:score}, by including the uncertainty term.
\begin{equation}
    S' = S - W_u \sum_{p \in CP'} \sigma_u,
    \label{eq:score}
\end{equation}
where $S$ is the grasp quality from \cite{pas_grasp_2017}, and the uncertainty value $\sigma_u$ corresponds to $CP'_{std}$ in Algorithm \ref{alg:grasp_alg}. The weight $W_u$ of the new term is adjusted depending on how much importance should be given to the uncertainty in the point cloud and on the grasp quality metric used. For GPD $W_u = 10^5$ since the standard deviation of the points are between 2-6 mm in absolute (non-normalized) values, and the scores from GPD are in hundreds or thousands of units, for PointNetGPD $W_u = 10^{-1}$ since the scores are from [0,1]. 
The last step (step 6) is then to reorder the grasp scores according to the new updated values. Figure \ref{fig:before_after_std} shows the order of the grasp transition to before running the uncertainty, to after the computation of grasp uncertainty. The colors of the grippers represent the ranking of the grasp candidates, from bright green as best, to bright red as worst score. It is important to not that we made the decision of generating the rank 15 grasp candidates to better evaluate the impact on the ranked 5 grasps. This choice did not affect the run-time of the whole pipeline since GPD is fairly optimized and running in a PC with 6 core, 12 threads CPU, 32 GB of RAM, and Nvidia GTX 1070 8 GB, generating 5, 10, or 15 grasps takes about 0.4-0.7 seconds with no significant difference between the three options. 

The algorithm \ref{alg:grasp_alg} provides, without changing the generation and classification of grasps, an update of the grasping score to account for the prediction error of shape completion networks. It was a familiarity decision to pick as the grasp pose generation algorithm GPD \cite{pas_grasp_2017} however, we also include the results of our pipeline with PointNetGPD \cite{liang_2019_pointnetgpd} as the grasp generation model.

%% file: 4_experiments.tex
\begin{table*}[h]
\caption{The grasp success rate for the ranked 5 best grasps candidates.}
\vspace{-0.25cm}
\centering
\resizebox{1\textwidth}{!}{%
\begin{tabular}{|l|c|c|c|c|c|c|c|c|c|c|c|}
\hline
 Method & Avg&  Pringles & Drill box &  Mustard & Mug & Cleanser &  Clamps (biggest) & Drill & Jell-o & Baseball & Pitcher with lid \\
\hline
rank 5 GPD \cite{pas_grasp_2017}  & $ 35\%$ & $ 42\%$&  $ 0\%$  &  $ 30\%$  & $ 30\%$ & $ 48\%$ & $ 34\%$ & $ 54\%$ & $ 54\%$ & $ 54\%$ & $ 2\%$ \\
\hline
rank 5 GPD+3DSGrasp \cite{mohammadi_3dsgrasp_2023} & $ \underline{51\%}$ & $ \underline{58\%}$&  $ \underline{38\%}$  &  $ \textbf{66\%}$  & $ 56\%$ & $ \underline{52\%}$ & $ \underline{48\%}$ & $ 54\%$ & $ \textbf{68\%}$ & $ 42\%$ & $ 26\%$ \\
\hline
rank 5 GPD+ours & $ \textbf{58\%}$ & $ \textbf{66\%}$&  $ \textbf{52\%}$  &  $ \underline{64\%}$  & $ 68\%$ & $ \textbf{72\%}$ & $ \textbf{52\%}$ & $ \textbf{64\%}$ & $ \underline{60\%}$ & $ \textbf{54\%}$ & $ \underline{28\%}$ \\
\hline
rank 5 PointNetGPD \cite{liang_2019_pointnetgpd}  & $ 18\%$ & $ 15\%$&  $ 2\%$  &  $ 5\%$  & $ 67\%$ & $ 22\%$ & $ 22\%$ & $ 16\%$ & $ 10\%$ & $ 0\%$ & $ 24\%$ \\
\hline
rank 5 PointNetGPD+3DSGrasp & $ 32\%$ & $ 15\%$&  $ 10\%$  &  $ 4\%$  & $ \textbf{80\%}$ & $ 30\%$ & $ 30\%$ & $ 46\%$ & $ 20\%$ & $ \underline{53\%}$ & $ 32\%$ \\
\hline
rank 5 PointNetGPD+ours & $ 49\%$ & $ 52\%$&  $ 33\%$  &  $ 52\%$  & $ \textbf{80\%}$ & $ 30\%$ & $ 33\%$ & $ \underline{63\%}$ & $ 40\%$ & $ 40\%$ & $ \textbf{40\%}$ \\
\hline
\end{tabular}}
\label{tab:grasp_sec_rate}
\end{table*}

\section{Robot Experiments}
\label{sec:robot}

In this section we present the robotic setup and the grasping experiments performed to evaluate our proposed approach against the state of the art method 3DSGrasp \cite{mohammadi_3dsgrasp_2023}. 

\noindent The robot platform used is the Kinova Gen3 and the gripper the Robotiq 2F-85. To mimic real-robot scenarios, the camera used is attached to the robot end-effector above the gripper, in this case of kinova gen3 it has already mounted an Intel RealSense D430 depth camera (visible in Figure \ref{fig:pipeline}). The robot scenario is a typical table-top where an object is placed on the table and the robot needs to find and grasp it successfully. This involves lifting and holding it for a reasonable time to confirm a stable grasp. In our experiments the robot is a 7-dof arm mounted on the side of the table and for a fair grasping metric the robot starts, in every grasp attempt, from a starting initial pose which views the table from a side-angle (as illustrated in top image from Figure \ref{fig:pipeline}). The initial pose aims to mimic a realistic view of how objects would be visible in household/human-centered environments, similar to Google DeepMind robots \cite{gu2023robotic}. In each trial, the robot starts from the initial position and an object is placed on the table at a random position and orientation, taking into account the reachability of the robot's arm. Using the camera, a single RGBD image is taken and converted as a PCD which after segmenting the table information gives us a single-view partial PCD of the object. From this we have the desired input for our robotic experiments. 

To run the shape completion network 3DSGrasp \cite{mohammadi_3dsgrasp_2023} we use the available code and pre-trained model to infer the completed PCD. Then, we generate the grasp candidates of the completed PCD using GPD \cite{pas_grasp_2017} or PointNetGPD \cite{liang_2019_pointnetgpd} as performed by their pipeline. To run our proposed pipeline we follow the same procedure as in 3DSGrasp with the addition of our uncertainty layer which instead of inferring once it infers 60 times to generate an average completed PCD and the standard deviation which is the uncertainty of our object shape. As mentioned in the previous section, we decided to evaluate the quality of the grasps for the ranked 5 from GPD and PointNetGPD, ranked 5 from 3DSGrasp and our proposed approach. In our approach we generate the top 15 and then, after running our grasp score algorithm, get the ranked 5 accounting for the uncertainty of the shape completion.
It is important to note that it is only provided the partial PCD, no background or table data are included to GPD and PointNetGPD. The robot platform, gripper, table-top setup, and initial position, and motion planner (MoveIt!), are identical for all methods for a fair comparison. 
The results are shown in Table \ref{tab:grasp_sec_rate}, with the average grasp success rate and the success rate for each object presented. We test on 10 YCB objects, each one for 10 trials, and in each trial we execute the rank 5 grasp candidates for all methods. There are 50 grasp attempts for each object making up for each method a total of 500 grasp candidates. In total the test experiments involved 3000 grasp candidates. Our pipeline takes on average 4 seconds to generate the completed PCD + uncertainty and 2 more seconds for running algorithm \ref{alg:grasp_alg}. There is an overall increase in the success rate when grasping objects using the new rank 5 grasp candidates compared to just using the rank 5 of GPD, GPD+3DSGrasp, PointNetGPD, and PointNetGPD+3DSGrasp (+23, +7, +31, +17 p.p., respectively). There are 2 objects present where our approach did not improve (mustard and jell-o) which could be a balancing issue between the weight of uncertainty compared to the original ranked scores. 

The ablation study presented in Table \ref{tab:ablation} compares our model with two other shape-completion models that also compute and measure uncertainty of object completion when grasping. It was not possible to reproduce their models so their results are the reported in their respective works. In addition, we assume that for each trial they only execute the rank 1 (highest ranked) grasp candidate. So as a comparison we present our results for rank 1 for the YCB objects present in all works. Our work achieves comparable performance even despite the fact that our uncertainty metric is only an additional parameter to the grasp quality metric.  


\begin{figure}[h]
    \centering
    \begin{tabular}{@{}c@{}}
        \centering
        \includegraphics[width=0.15\textwidth,height=0.15\textheight]{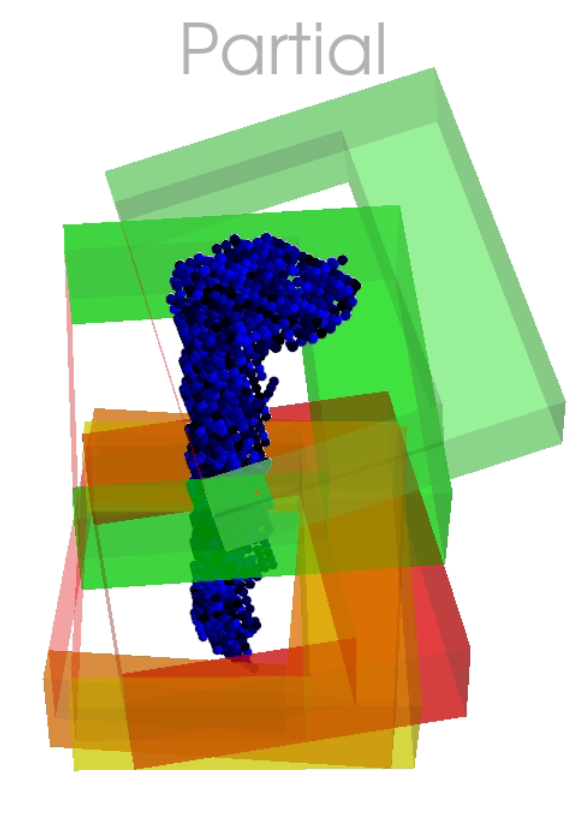}
    \end{tabular}
    \begin{tabular}{@{}c@{}}
        \centering 
        \includegraphics[width=0.15\textwidth,height=0.15\textheight, trim={0 0 0 2cm}]{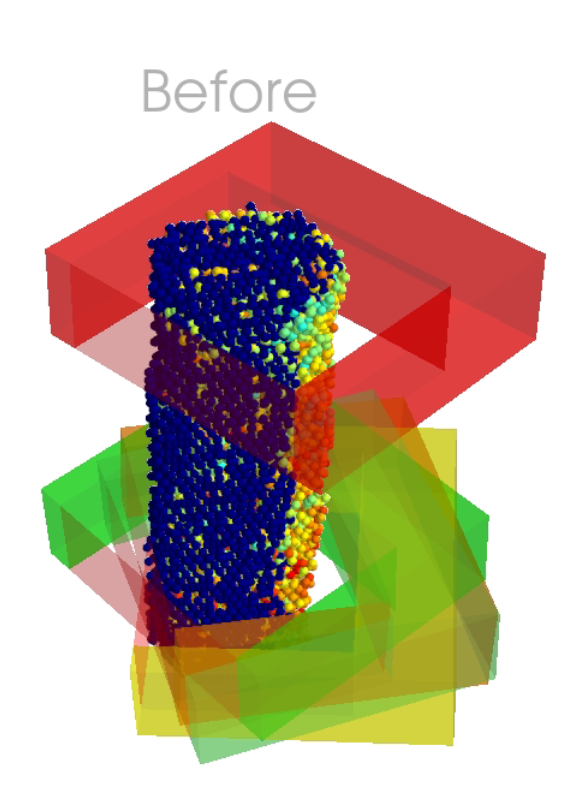}
    \end{tabular}
    \begin{tabular}{@{}c@{}}
        \centering 
        \includegraphics[width=0.15\textwidth,height=0.15\textheight, trim={0 0 0 1cm}]{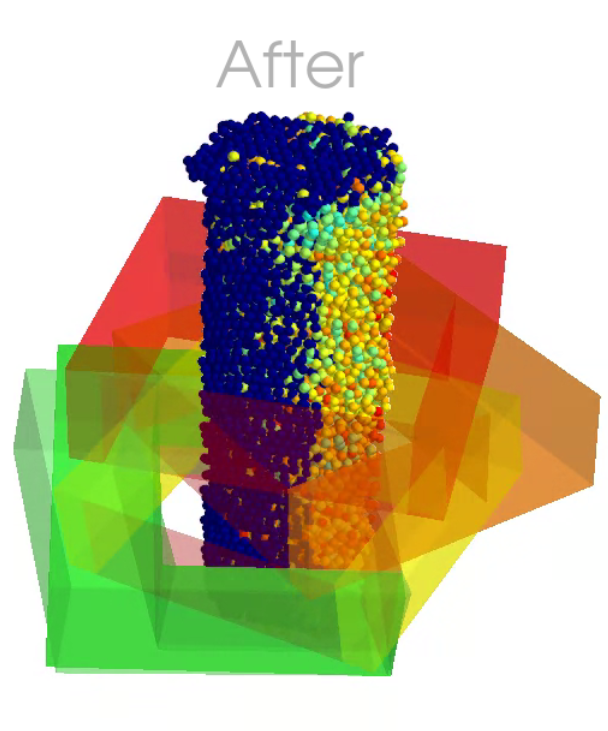}
    \end{tabular}
    \caption{The rank 5 grasp candidates for each of the three methods: GPD, 3DSGrasp, and ours. The GPD method often leads to collisions and/or grasps that are unreachable by the robot; 3DSGrasp solves the issue with object collisions but still has many unreachable grasps; our method keeps the benefits of 3DSGrasp but produces less unreachable grasps.}
    \label{fig:exp}
\end{figure}

\begin{table}[h]
\caption{Ablation Study of Shape-Completion-Based Grasping with Uncertainty}
\centering
\resizebox{0.49\textwidth}{!}{%
\begin{tabular}{|l|c|c|c|c|}
\hline
 Method & Drill & Mustard & Drill box &  Pitcher with lid \\
\hline
rank 1 USN \cite{lundell2019robust}  & $ 70\%$ & $ \textbf{90\%}$&  $ \textbf{80\%}$  &  $ 30\%$ \\
\hline
rank 1 HyperPCR \cite{rosasco2022towards} & $ \textbf{100\%}$ & $ \textbf{90\%}$&  $ \textbf{80\%}$  &  $ 20\%$ \\
\hline
rank 1 GPD+ours & $ \underline{80\%}$ & $ \textbf{90\%}$&  $ \textbf{80\%}$  &  $ \textbf{40\%}$ \\
\hline
\end{tabular}}
\label{tab:ablation}
\end{table}

\section{Robot Experiments}
\label{sec:robot}

In Figure \ref{fig:exp} it illustrates one of the key benefits of using our approach. GPD \cite{pas_grasp_2017} has been used with great results for top-down grasps on a plethora of objects, however for side views where self-occlusions are present it can generate grasps that collide with the object (as shown in the ``Partial'' solution from Figure \ref{fig:exp}). The introduction of shape completion models (like 3DSGrasp) solve this issue but a new one arises, grasps on the opposite direction of the robot (as shown in the ``Before'' solution from Figure \ref{fig:exp}). GPD simply processes the completed PCD without any knowledge of what are real points (partial) and generated points (complete). Our approach provides the best of both worlds by including the benefits of having real points (zero uncertainty) and the generated completed shape (with uncertainty) to penalize grasps that are further away from the camera's point of view. This is not only a more robust grasp quality score because it accounts for the noise of model's output but at the same time it favours the visible regions of the object that is also the side that is closest to the robot. 


%% file: 5_conclusion.tex
\section{Conclusion} 
\label{sec:conclusion}

In this work we present a grasping pipeline that improves on previous 3D single-view object shape completion approaches \cite{mohammadi_3dsgrasp_2023} by measuring and adding the error of the object shape prediction in the decision step for the grasp candidates. The grasp quality score is the metric to rank from the generated grasp candidates which have the best chance of resulting in successful (good) grasps. Current off-the-shelf grasp generation algorithms \cite{pas_grasp_2017, liang_2019_pointnetgpd} have learned to generate good grasps based on force-closure grasp metrics or evaluating based on the points inside the gripper. Our approach, presented in Section \ref{sec:grasp}, computes the uncertainty of each grasp and adds this factor to the quality score to update the ranking of all grasp candidates. The grasp uncertainty is approximated by the Monte Carlo Dropout technique, presented in Section \ref{sec:model}, which measures the shape uncertainty of the completed PCD. 
The extension of state of the art vision-based completion for grasping pipelines \cite{mohammadi_3dsgrasp_2023} with the inclusion of completion uncertainty into the grasping quality leads to more accurate assessment of grasp successfulness in relation to ranked grasp quality score. 

The experiments evaluated the grasp success rate of the top 5 grasps proposed by each of the pipelines: the partial point cloud (GPD \cite{pas_grasp_2017} or PointNetGPD \cite{liang_2019_pointnetgpd}), a shape completion network (3DSGrasp \cite{mohammadi_3dsgrasp_2023}), and our completion uncertainty approach. We can conclude that accounting for the uncertainty of shape completion of objects in grasping experiments outperforms all the other approaches in providing more robust grasp scores for a higher success rate. 
For future work, we plan to develop a grasp generator that takes into account the uncertainty of shape completion models. 


%% file: 0_root.bbl
\begin{thebibliography}{10}

\bibitem{han2017high}
X.~Han, Z.~Li, H.~Huang, E.~Kalogerakis, and Y.~Yu, ``\href{https://openaccess.thecvf.com/content_iccv_2017/html/Han_High-Resolution_Shape_Completion_ICCV_2017_paper.html}{High-resolution shape completion using deep neural networks for global structure and local geometry inference},'' in {\em Proceedings of the IEEE international conference on computer vision}, pp.~85--93, 2017.

\bibitem{mohammadi2022svp}
S.~S. Mohammadi, Y.~Wang, M.~Taiana, P.~Morerio, and A.~Del~Bue, ``\href{https://link.springer.com/chapter/10.1007/978-3-031-06430-2_2}{SVP-Classifier: Single-View Point Cloud Data Classifier with Multi-view Hallucination},'' in {\em International Conference on Image Analysis and Processing}, pp.~15--26, Springer, 2022.

\bibitem{varley2017shape}
J.~Varley, C.~DeChant, A.~Richardson, J.~Ruales, and P.~Allen, ``\href{https://ieeexplore.ieee.org/abstract/document/8206060}{Shape completion enabled robotic grasping},'' in {\em 2017 IEEE/RSJ international conference on intelligent robots and systems (IROS)}, pp.~2442--2447, IEEE, 2017.

\bibitem{mohammadi_3dsgrasp_2023}
S.~S. Mohammadi, N.~F. Duarte, D.~Dimou, Y.~Wang, M.~Taiana, P.~Morerio, A.~Dehban, P.~Moreno, A.~Bernardino, A.~Del~Bue, and J.~Santos-Victor, ``\href{https://ieeexplore.ieee.org/document/10160350}{3DSGrasp: 3D Shape-Completion for Robotic Grasp},'' in {\em 2023 {IEEE} {International} {Conference} on {Robotics} and {Automation} ({ICRA})}, pp.~3815--3822, 2023.

\bibitem{pas_grasp_2017}
A.~ten Pas, M.~Gualtieri, K.~Saenko, and R.~Platt, ``\href{https://arxiv.org/abs/1706.09911}{Grasp Pose Detection in Point Clouds},'' {\em The International Journal of Robotics Research}, vol.~36, no.~13-14, pp.~1455--1473, 2017.

\bibitem{fang2020graspnet}
H.-S. Fang, C.~Wang, M.~Gou, and C.~Lu, ``\href{https://openaccess.thecvf.com/content_CVPR_2020/papers/Fang_GraspNet-1Billion_A_Large-Scale_Benchmark_for_General_Object_Grasping_CVPR_2020_paper.pdf}{Graspnet-1billion: A large-scale benchmark for general object grasping},'' in {\em Proceedings of the IEEE/CVF conference on computer vision and pattern recognition}, pp.~11444--11453, 2020.

\bibitem{fang_anygrasp_2023}
H.-S. Fang, C.~Wang, H.~Fang, M.~Gou, J.~Liu, H.~Yan, W.~Liu, Y.~Xie, and C.~Lu, ``\href{https://ieeexplore.ieee.org/document/10167687}{AnyGrasp: Robust and Efficient Grasp Perception in Spatial and Temporal Domains},'' vol.~39, pp.~3929--3945, 2023.

\bibitem{berger2014state}
M.~Berger, A.~Tagliasacchi, L.~Seversky, P.~Alliez, J.~Levine, A.~Sharf, and C.~Silva, ``\href{https://infoscience.epfl.ch/record/299156}{State of the art in surface reconstruction from point clouds},'' {\em Eurographics 2014-State of the Art Reports}, vol.~1, no.~1, pp.~161--185, 2014.

\bibitem{de2015efficient}
R.~P. de~Figueiredo, P.~Moreno, and A.~Bernardino, ``\href{https://www.sciencedirect.com/science/article/abs/pii/S092523121401265X}{Efficient pose estimation of rotationally symmetric objects},'' {\em Neurocomputing}, vol.~150, pp.~126--135, 2015.

\bibitem{kazhdan2006poisson}
M.~Kazhdan, M.~Bolitho, and H.~Hoppe, ``\href{https://hhoppe.com/poissonrecon.pdf}{Poisson surface reconstruction},'' in {\em Proceedings of the fourth Eurographics symposium on Geometry processing}, vol.~7, p.~0, 2006.

\bibitem{gurumurthy2019high}
S.~Gurumurthy and S.~Agrawal, ``\href{https://ieeexplore.ieee.org/abstract/document/8658987}{High fidelity semantic shape completion for point clouds using latent optimization},'' in {\em 2019 IEEE Winter Conference on Applications of Computer Vision (WACV)}, pp.~1099--1108, IEEE, 2019.

\bibitem{qi2017pointnet}
C.~R. Qi, L.~Yi, H.~Su, and L.~J. Guibas, ``\href{https://proceedings.neurips.cc/paper_files/paper/2017/hash/d8bf84be3800d12f74d8b05e9b89836f-Abstract.html}{Pointnet++: Deep hierarchical feature learning on point sets in a metric space},'' {\em Advances in neural information processing systems}, vol.~30, 2017.

\bibitem{yuan2018pcn}
W.~Yuan, T.~Khot, D.~Held, C.~Mertz, and M.~Hebert, ``\href{https://ieeexplore.ieee.org/abstract/document/8491026}{Pcn: Point completion network},'' in {\em 2018 International Conference on 3D Vision (3DV)}, pp.~728--737, IEEE, 2018.

\bibitem{yang2018foldingnet}
Y.~Yang, C.~Feng, Y.~Shen, and D.~Tian, ``\href{https://openaccess.thecvf.com/content_cvpr_2018/html/Yang_FoldingNet_Point_Cloud_CVPR_2018_paper.html}{Foldingnet: Point cloud auto-encoder via deep grid deformation},'' in {\em Proceedings of the IEEE Conf. on computer vision and pattern recognition}, pp.~206--215, 2018.

\bibitem{groueix2018papier}
T.~Groueix, M.~Fisher, V.~G. Kim, B.~C. Russell, and M.~Aubry, ``\href{https://openaccess.thecvf.com/content_cvpr_2018/html/Groueix_A_Papier-Mache_Approach_CVPR_2018_paper.html}{A papier-m{\^a}ch{\'e} approach to learning 3d surface generation},'' in {\em Proceedings of the IEEE conference on computer vision and pattern recognition}, pp.~216--224, 2018.

\bibitem{liu2020morphing}
M.~Liu, L.~Sheng, S.~Yang, J.~Shao, and S.-M. Hu, ``\href{https://ojs.aaai.org/index.php/AAAI/article/view/6827}{Morphing and sampling network for dense point cloud completion},'' in {\em Proceedings of the AAAI conference on artificial intelligence}, vol.~34, pp.~11596--11603, 2020.

\bibitem{pan2021multi}
L.~Pan, T.~Wu, Z.~Cai, Z.~Liu, X.~Yu, Y.~Rao, J.~Lu, J.~Zhou, M.~Xu, X.~Luo, {\em et~al.}, ``\href{https://arxiv.org/abs/2112.12053}{Multi-View Partial (MVP) Point Cloud Challenge 2021 on Completion and Registration: Methods and Results},'' {\em arXiv preprint arXiv:2112.12053}, 2021.

\bibitem{yu2021PoinTr}
X.~Yu, Y.~Rao, Z.~Wang, Z.~Liu, J.~Lu, and J.~Zhou, ``\href{https://openaccess.thecvf.com/content/ICCV2021/html/Yu_PoinTr_Diverse_Point_Cloud_Completion_With_Geometry-Aware_Transformers_ICCV_2021_paper.html}{Pointr: Diverse point cloud completion with geometry-aware transformers},'' in {\em Proceedings of the IEEE/CVF international conference on computer vision}, pp.~12498--12507, 2021.

\bibitem{chen2022improving}
W.~Chen, H.~Liang, Z.~Chen, F.~Sun, and J.~Zhang, ``\href{https://link.springer.com/article/10.1007/s10846-022-01586-4}{Improving Object Grasp Performance via Transformer-Based Sparse Shape Completion},'' {\em Journal of Intelligent \& Robotic Systems}, vol.~104, no.~3, pp.~1--14, 2022.

\bibitem{PointSDFgrasping2020}
M.~Van~der Merwe, Q.~Lu, B.~Sundaralingam, M.~Matak, and T.~Hermans, ``\href{https://ieeexplore.ieee.org/abstract/document/9196981}{Learning Continuous 3D Reconstructions for Geometrically Aware Grasping},'' in {\em 2020 IEEE International Conference on Robotics and Automation (ICRA)}, pp.~11516--11522, 2020.

\bibitem{yang2021robotic}
D.~Yang, T.~Tosun, B.~Eisner, V.~Isler, and D.~Lee, ``\href{https://ieeexplore.ieee.org/abstract/document/9562046}{Robotic grasping through combined image-based grasp proposal and 3d reconstruction},'' in {\em 2021 IEEE International Conference on Robotics and Automation (ICRA)}, pp.~6350--6356, IEEE, 2021.

\bibitem{humt_combining_2023}
M.~Humt, D.~Winkelbauer, U.~Hillenbrand, and B.~Bäuml, ``\href{https://ieeexplore.ieee.org/document/10375210/}{Combining Shape Completion and Grasp Prediction for Fast and Versatile Grasping with a Multi-Fingered Hand},'' in {\em 2023 {IEEE}-{RAS} 22nd {International} {Conference} on {Humanoid} {Robots} ({Humanoids})}, (Austin, TX, USA), pp.~1--8, IEEE, Dec. 2023.

\bibitem{li2016dexterous}
M.~Li, K.~Hang, D.~Kragic, and A.~Billard, ``\href{https://www.sciencedirect.com/science/article/abs/pii/S0921889015001967}{Dexterous grasping under shape uncertainty},'' {\em Robotics and Autonomous Systems}, vol.~75, pp.~352--364, 2016.

\bibitem{rosasco2022towards}
A.~Rosasco, S.~Berti, F.~Bottarel, M.~Colledanchise, and L.~Natale, ``\href{https://arxiv.org/pdf/2209.04300}{Towards confidence-guided shape completion for robotic applications},'' in {\em 2022 IEEE-RAS 21st International Conference on Humanoid Robots (Humanoids)}, pp.~580--586, IEEE, 2022.

\bibitem{chen2018probabilistic}
D.~Chen, V.~Dietrich, Z.~Liu, and G.~Von~Wichert, ``\href{https://link.springer.com/article/10.1007/s10846-017-0646-y}{A probabilistic framework for uncertainty-aware high-accuracy precision grasping of unknown objects},'' {\em Journal of Intelligent \& Robotic Systems}, vol.~90, pp.~19--43, 2018.

\bibitem{lundell2019robust}
J.~Lundell, F.~Verdoja, and V.~Kyrki, ``\href{https://ieeexplore.ieee.org/abstract/document/8967816}{Robust grasp planning over uncertain shape completions},'' in {\em 2019 IEEE/RSJ International Conference on Intelligent Robots and Systems (IROS)}, pp.~1526--1532, IEEE, 2019.

\bibitem{rustler_active_2022}
L.~Rustler, J.~Lundell, J.~K. Behrens, V.~Kyrki, and M.~Hoffmann, ``\href{https://ieeexplore.ieee.org/document/9720238}{Active Visuo-Haptic Object Shape Completion},'' {\em IEEE Robotics and Automation Letters}, vol.~7, no.~2, pp.~5254--5261, 2022.

\bibitem{rustler_efficient_2023}
L.~Rustler, J.~Matas, and M.~Hoffmann, ``\href{https://ieeexplore.ieee.org/document/10342200/}{Efficient Visuo-Haptic Object Shape Completion for Robot Manipulation},'' in {\em 2023 {IEEE}/{RSJ} {International} {Conference} on {Intelligent} {Robots} and {Systems} ({IROS})}, (Detroit, MI, USA), pp.~3121--3128, IEEE, Oct. 2023.

\bibitem{humt2023shape}
M.~Humt, D.~Winkelbauer, and U.~Hillenbrand, ``\href{https://arxiv.org/pdf/2308.00377.pdf}{Shape completion with prediction of uncertain regions},'' in {\em 2023 IEEE/RSJ International Conference on Intelligent Robots and Systems (IROS)}, pp.~1215--1221, IEEE, 2023.

\bibitem{guo2021pct}
M.-H. Guo, J.-X. Cai, Z.-N. Liu, T.-J. Mu, R.~R. Martin, and S.-M. Hu, ``\href{https://link.springer.com/article/10.1007/s41095-021-0229-5}{Pct: Point cloud transformer},'' {\em Computational Visual Media}, vol.~7, pp.~187--199, 2021.

\bibitem{pmlr-v48-gal16}
Y.~Gal and Z.~Ghahramani, ``\href{https://proceedings.mlr.press/v48/gal16.html}{Dropout as a Bayesian Approximation: Representing Model Uncertainty in Deep Learning},'' in {\em Proceedings of The 33rd International Conference on Machine Learning} (M.~F. Balcan and K.~Q. Weinberger, eds.), vol.~48 of {\em Proceedings of Machine Learning Research}, (New York, New York, USA), pp.~1050--1059, PMLR, 20--22 Jun 2016.

\bibitem{liang_2019_pointnetgpd}
H.~Liang, X.~Ma, S.~Li, M.~Görner, S.~Tang, B.~Fang, F.~Sun, and J.~Zhang, ``\href{https://ieeexplore.ieee.org/document/8794435}{PointNetGPD: Detecting Grasp Configurations from Point Sets},'' pp.~3629--3635, 2019.

\bibitem{gu2023robotic}
J.~Gu, S.~Kirmani, P.~Wohlhart, Y.~Lu, M.~G. Arenas, K.~Rao, W.~Yu, C.~Fu, K.~Gopalakrishnan, Z.~Xu, {\em et~al.}, ``\href{https://arxiv.org/abs/2311.01977}{Robotic Task Generalization via Hindsight Trajectory Sketches},'' in {\em First Workshop on Out-of-Distribution Generalization in Robotics at CoRL 2023}, 2023.

\end{thebibliography}
